 \documentclass[pmlr,twocolumn,10pt,preprint]{jmlr} 

\usepackage{booktabs}
\usepackage{verbatim}
\usepackage{siunitx}
\usepackage[switch]{lineno}
\usepackage{multirow}
\usepackage{tikz}
\usepackage{pifont}

\theorembodyfont{\upshape}
\theoremheaderfont{\scshape}
\theorempostheader{:}
\theoremsep{\newline}

\setcitestyle{numbers,square}

 \title{A Multimodal Data Processing Pipeline for MIMIC-IV Dataset}

\author{%
\Name{Farzana Islam Adiba} \Email{fadiba@udel.edu}
\AND
\Name{Varsha Danduri}\Email{varshada@udel.edu}
\AND
\Name{Fahmida Liza Piya}\Email{lizapiya@udel.edu}
\AND
\Name{Ali Abbasi}\Email{aiai@udel.edu}
\AND
\Name{Mehak Gupta}\Email{mehakg@udel.edu}
\AND
\Name{Rahmatollah Beheshti}\Email{rbi@udel.edu}
\AND
\small
\begin{center}
University of Delaware 
\end{center}
}

\begin{document}

\maketitle
\thispagestyle{plain}
\begin{abstract}
The MIMIC-IV dataset is a large, publicly available electronic health record (EHR) resource widely used for clinical machine learning research. It comprises multiple modalities, including structured data, clinical notes, waveforms, and imaging data. Working with these disjointed modalities requires an extensive manual effort to preprocess and align them for downstream analysis. While several pipelines for MIMIC-IV data extraction are available, they target a small subset of modalities or do not fully support arbitrary downstream applications.  In this work, we greatly expand our prior popular unimodal pipeline and present a comprehensive and customizable multimodal pipeline that can significantly reduce multimodal processing time and enhance the reproducibility of MIMIC-based studies. Our pipeline systematically integrates the listed modalities, enabling automated cohort selection, temporal alignment across modalities, and standardized multimodal output formats suitable for arbitrary static and time-series downstream applications. We release the code, a simple UI, and a Python package for selective integration (with embedding) at \url{https://github.com/healthylaife/MIMIC-IV-Data-Pipeline}.
\end{abstract}



\section{Introduction}
\label{sec:intro}
The MIMIC dataset is the most widely used electronic health record (EHR) dataset in various research studies. The comprehensive, open-source nature of the dataset, along with its diverse range of populations, including various disease cohorts for both ICU and non-ICU patients, makes the dataset highly adaptable to the diverse needs of the research community \citep{johnson2023mimic}. MIMIC-IV (the latest available version of MIMIC) comprises two major modules: the hospital (hosp) and ICU (icu) modules, which encompass vital signs, laboratory results, medication orders, diagnoses, procedures, and demographic information \citep{johnson2024mimic3}. The unstructured component (note module) offers rich clinical narratives such as discharge summaries and radiology reports.

Over time, the MIMIC data have also been enriched with additional modalities beyond structured and unstructured datasets, including waveforms and imaging. MIMIC-CXR, a companion dataset linked to MIMIC-IV, contains chest X-ray images paired with corresponding radiology reports, enabling image-text research in clinical diagnostics. The 2D Echocardiogram images include ultrasound images offering a sequential view of the heart \citep{gow2023mimicecho}. MIMIC-IV also includes waveforms of patients' physiological or electrocardiogram (ECG) signals \citep{wfdb2022, mimicwaveforms, gow2023mimic-ecg}. These heterogeneous modalities offer crucial resources for various artificial intelligence (AI) models that aim to accurately capture the complex and multimodal nature of real-world healthcare. 

However, these resources are distributed across separate modules and stored independently, with no unified pipeline provided for seamless integration. As a result, researchers are often required to perform extensive manual preprocessing, temporal alignment, and feature engineering to unify these data sources. This separation not only introduces extra work and inconsistency in preprocessing workflows but also makes reproducibility and comparability across studies difficult.

While prior work has attempted to address the preprocessing and benchmarking of MIMIC datasets, they often focus on a single modality \citep{wang2020mimicextract, johnson2019mimic-cxr}. Several studies have been conducted on MIMIC's multimodal application; however, they also have various limitations. These limitations include, i) their primary objectives being to perform downstream tasks rather than a generalizable pipeline for obtaining integrated modalities (e.g., \citep{golovanevsky2025picme}), ii) being limited to a smaller subset of modalities (e.g., \texttt{MIMIC-Extract} \citep{wang2020mimicextract} and \texttt{COP-E-CAT}\citep{mandyam2021cop}), and iii) not fully supporting customizable extraction or arbitrary downstream tasks controlled by the end-user (e.g., \citep{soenksen2022haim}).

To address the above gap, we present a multimodal pipeline that integrates the five major modalities in the MIMIC-IV dataset. By merging these into a single, integrated DataFrame, our pipeline aims to substantially lower the entry barrier for researchers working with the MIMIC family of datasets and democratize easy access to multimodal EHR research, enabling both experienced and early-stage researchers to focus more directly on scientific questions rather than technical infrastructure. It also aims to reduce the errors that may impact downstream usage of data across different applications. In particular, the major contributions of this work are:
\begin{itemize}
    \item We develop an extensive multimodal pipeline for MIMIC-IV, enabling the raw integration of structured EHR data, unstructured clinical notes, waveforms, X-ray images, and 2-D echocardiograms into a unified framework suitable for downstream tasks.
    \item We extend the integration by generating normalized embeddings for each modality, assisting users to directly use these for modeling experiments.
    \item Our pipeline supports flexible cohort selection based on the International Classification of Diseases (ICD) codes, cross-modality mapping, and sectionizing clinical notes. 
\end{itemize}
\section{Related Works}\label{sec:related_work}

\paragraph{Unimodal Pipelines.}
\label{sec:nonmodal}
A variety of data processing pipelines with MIMIC datasets have been developed over the years, with many MIMIC-III pipelines primarily focusing on the unimodal nature of the framework. Notably, \citet{harutyunyan2019multitask} and \citet{wang2020mimicextract} developed a comprehensive framework for this dataset to perform downstream tasks using ML methods for prediction or classification. The former one was mainly built to process clinical events based on ICU data, and the latter developed a cohort of 34,472 patients with diverse demographic and admission coverage, which has been employed in ML experiments. \texttt{FIDDLE} (Flexible Data-Driven Pipeline) is another flexible preprocessing pipeline, where they used eICU data apart from MIMIC-III to support extracting patient cohorts with largely automated feature engineering steps \citep{tang2020fiddle}. 

The release of the MIMIC-IV has opened new doorways, providing expanded, comprehensive patient information, broadening the applicability of the MIMIC-based pipelines. Adapting this new addition, a cohort selection pipeline, \texttt{COP-E-CAT} \citep{mandyam2021cop}, with feature engineering features such as noise removal, missing values imputation, was introduced. Another extensive pipeline was introduced by \citet{gupta2022extensive}, where our team offered adaptability for patient groups by supporting outlier removal and imputation for predictive modeling, while incorporating fairness evaluations. Other pipelines, such as the Flexible Pipeline for Time-Series Data \citep{chen4737942timeseriesmimic}, targeted the latest MIMIC-IV and MIMIC-IV-ED \citep{johnson2021mimic-ed} versions, and built visualization tools for exploratory analysis. \texttt{METRE} (Multidatabase ExTRaction Pipeline) \citep{liao2023metre} is another unimodal work that used eICU along with MIMIC-IV, representing an advancement by supporting consistent data extraction through cross-database modeling and comparative validation.

Although these structured pipelines provided useful methods for establishing automated and reproducible data preprocessing pipelines, the structured characteristics of these pipelines limited their applicability to complex, multimodal clinical tasks. Our work focuses on mitigating this gap, enhancing the multimodal integration of the MIMIC-IV dataset. 

\paragraph{Multi-modal Pipelines.}
\label{sec:multimodal}
Several studies have implemented multi-modal features from MIMIC datasets. \texttt{AutoFM} \citep{cui2024autofm} is an extended work of \texttt{FIDDLE} \citep{tang2020fiddle} that processed the discrete and continuous events from MIMIC-III as separate time-series input modalities, which were processed via neural architectural search (NAS) for predicting diagnosis codes and multi-label classification for discharge-based prediction. The input modalities implemented in \texttt{AutoFM} are unstructured notes, discrete events (e.g., medical codes), and continuous events (e.g., ICU lab reports). Similarly, \texttt{Quick-MIMIC} by \citet{dou2024quickmimic} built a data extraction pipeline analyzing time-series data with structured and unstructured data. Apart from these modalities, \texttt{MDS-ED} \citep{alcaraz2025mds-ed} additionally included ECG waveforms from the MIMIC database \citep{gow2023mimic-ecg} and integrated this pipeline for task-specific prediction. \texttt{PulseDB} \citep{wang2023pulsedb} and \texttt{MIMIC-BP} \citep{sanches2024mimicbp} both of these pipelines used the MIMIC-III Waveform dataset \citep{moody2020mimic3wdb}, consisting of ECG, photoplethysmography (PPG), and arterial BP (ABP) waveforms for PulseDB, while MIMIC-BP additionally used respiration (RESP) for analysis with ML models. \texttt{MIMIC-IV-Ext-22MCTS} \citep{wangmimicivext} is a new edition for time-series datasets from discharge summaries from the MIMIC-IV-Note by chunking the texts with fine-tuned BERT models, while added prompts to identify the related chunks for specific clinical events.  

Most of these pipelines are developed with a focus on specific tasks, such as time-series prediction or diagnosis classification, emphasizing either text or signals, rather than imaging. On the other hand, \texttt{MEETI} \citep{zhang2025meeti}, \texttt{Symile-MIMIC} \citep{saportasymile}, and \texttt{PicME} \citep{golovanevsky2025picme} all used the MIMIC-CXR dataset \citep{johnson2019mimic-cxr} in their pipeline for integrating image modalities. \texttt{MIMIC-Eye} \citep{hsieh2023mimic-eye} and \texttt{REFLACX} \citep{lanfredi2021reflacx} are both based on the MIMIC-IV-CXR dataset, but mainly collect eyetracking data of radiologists while analyzing the CXR images. \citet{soenksen2022haim} introduced another notable multi-modal framework, \texttt{HAIM-MIMIC-MM}, with MIMIC-IV and MIMIC-CXR-JPG, considering multiple modalities as text, structured data with timeseries data, and radiology images. Their major contribution was to prepare different input modalities of information that are concatenated into a fixed fusion embedding for downstream predictive analysis. \citet{yu2025mimic-multimodal} constructed a benchmark ICU-stay dataset on MIMIC-IV following the HAIM framework, integrating Hospital, ICU, CXR, and Notes modules to benchmark unimodal and multimodal concatenated models' embeddings on mortality and LOS tasks. However, this work did not focus on developing a customizable patient cohort selection approach, which is crucial for domain-specific research. Our work addresses these gaps by focusing on enhancing customized cohort selection while aiming to further reduce the time and memory constraints associated with large-scale multimodal data processing.











\section{Method}\label{sec:design}
\begin{figure*}[ht!]
    \centering
    \includegraphics[width=0.9\linewidth]{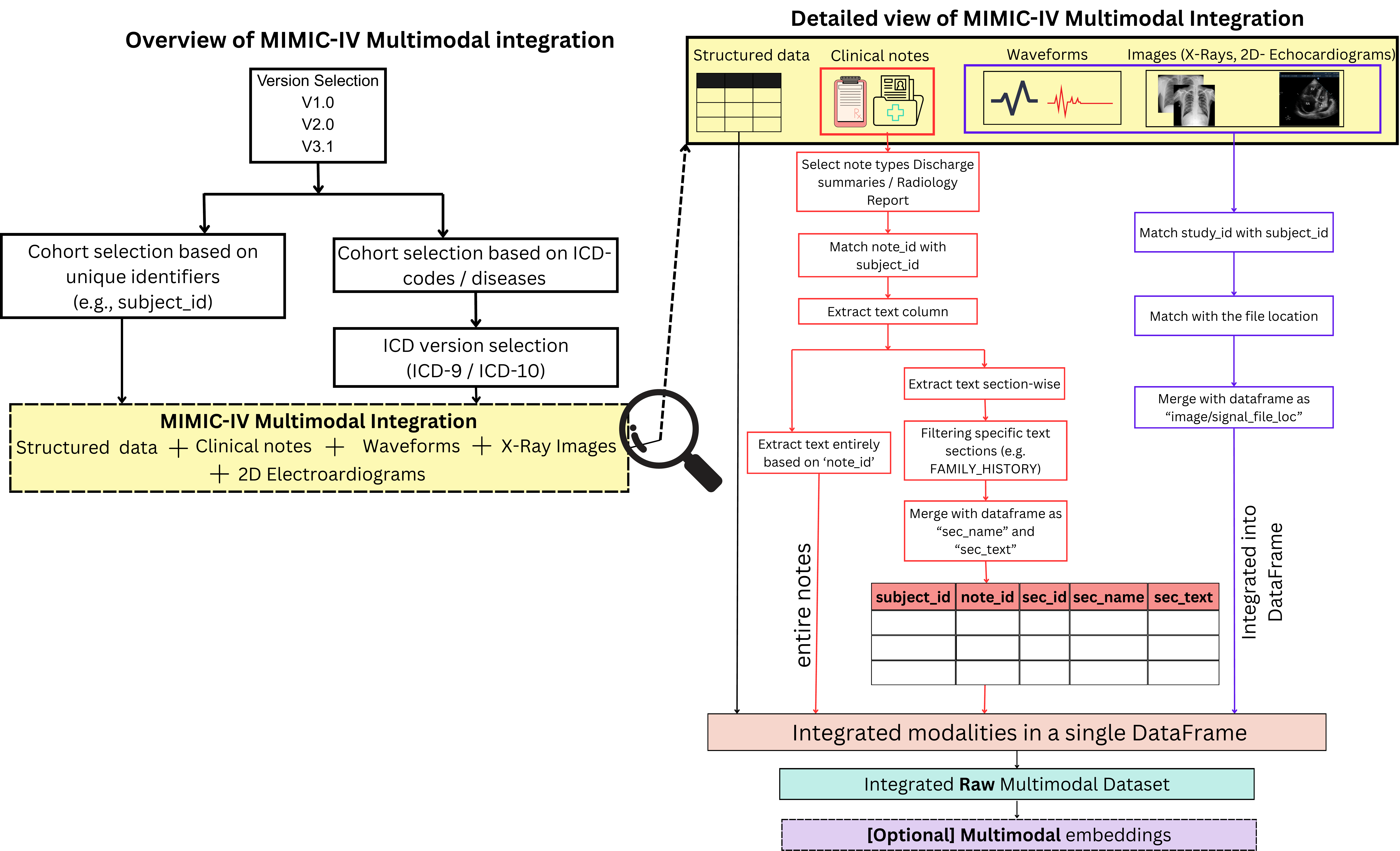}
    \caption{The overview of the proposed MIMIC-IV multimodal pipeline. The right part shows a zoomed-in version of the yellow box on the left.}
    \label{fig:framework}
\end{figure*}

We present a multimodal pipeline that unifies structured EHR data, clinical notes, waveforms, X-ray images, and 2D-echocardiograms 
using stable identifier spines (\texttt{subject\_id} for patients, \texttt{hadm\_id} for encounters, \texttt{note\_id} for notes, and \texttt{study\_id} for imaging). As shown on the top left of Figure \ref{fig:framework}, the system supports two levels of integration: i) multimodal integration via identifiers, which enumerates all modalities, linked to patients/encounters across the corpus; and ii) ICD-based cohorting, which filters patients and admissions by ICD-9/10 codes and then retrieves only their related information across the modalities.  

\subsection{Datasets}
Our pipeline applies to MIMIC-IV \citep{johnson2024mimic31}, which is an extensive, publicly available, de-identified EHR resource from Beth Israel Deaconess Hospital, from a collection of patient admissions between the years 2008 and 2019. MIMIC-IV is divided into two main modules, `hosp' and `icu,' consisting of patient data for over a decade of admissions. 

Our pipeline covers five types of resources from the MIMIC family of datasets, each representing a distinct clinical modality. MIMIC-IV provides the structured and time-series format of EHRs, capturing demographics, diagnoses, laboratory results, and ICU stays over more than a decade of admissions. 

On the other hand, MIMIC-IV-Note provides rich narrative documentation, including discharge summaries and radiology reports, which capture clinical histories, patient background, and image interpretations in free-text form. Additionally, MIMIC-ECG provides waveform recordings of electrocardiograms, along with study-level metadata, including cardiovascular signals. Other types of waveform data are also present in the MIMIC-IV Waveform Database. MIMIC-CXR-JPG adds chest radiographs with standardized metadata, enabling direct linkage between imaging, text, and structured cohorts. Lastly, the MIMIC-IV-ECHO dataset contains 2D Echocardiograms. 


\subsection{Pipeline elements}
\subsubsection{Data Filtering and Cohort Selection}\label{dat_filter}
Our wizard-style pipeline begins with the selection of options for the different MIMIC-IV versions (currently, versions 1, 2, and 3.1). As mentioned earlier, a user can choose to use the shared unique identifiers as \texttt{subject\_id} to integrate different modalities based on the matching identifiers. 
%
%
%
Alternatively, the user can opt for an ICD-based cohort creation that is connected based on the \texttt{diagnoses\_icd.csv} file from the \texttt{hosp} module. This path would have options for selecting both ICD-9 and ICD-10 versions, and the pipeline gives further flexibility as users can filter by exact ICD codes or by their associated disease names. 
The ICD path offers a choice between fine-grained ICD code-level analysis and broader disease-category studies. 

\subsubsection{Structured Data Processing} \label{sec:tab-data}

The structured (tabular) data from the \texttt{hosp} and \texttt{icu} modules includes admission information, lab events, and chart events, along with the admission time of the patients. These timestamps are used for temporal handling and analysis. The \texttt{subject\_id} serves as a unique identifier for each patient and is used for bridging to other forms of information, including text, images, and signals. 


\subsubsection{Clinical notes processing}\label{sec:note-process}
The pipeline currently uses MIMIC-IV-Note for text type integration, which includes free-text clinical documentation associated with MIMIC-IV encounters, including discharge summaries (DS) and radiology reports (RR) that cover common clinical workflows. Each note carries a unique \texttt{note\_id} and links to \texttt{subject\_id}. Most of the inpatient notes also have a \texttt{hadm\_id} for each \texttt{subject\_id}. 

The pipeline uses a section segmentation approach that extracts sections from the available text. It employs a rule-based approach using regular expressions to identify and isolate specific sections related to patient history. The pipeline uses regex-based feature extraction tools to retrieve raw text based on sections related to patients' history, which included \texttt{CHIEF COMPLAINT}, \texttt{HISTORY OF PRESENT ILLNESS}, \texttt{PAST MEDICAL HISTORY}, and \texttt{SOCIAL HISTORY}. The sectionizer identifies section headers for the matched text spans and assigns them to a section header until a new header is encountered. The matched patterns are derived from standard clinical note conventions (e.g., ``Clinical Assessment:” or ``Past Medical History:”) and can be customized to include other (e.g., institution-specific) variants. 
By integrating this component into the section extraction tools, the notes are parsed across the entire corpus.

Aiming to create a DataFrame (2D tabular structure), each extracted note is then represented as a collection of subsections as (\texttt{section\_id}, \texttt{section\_name}, \texttt{section\_text}), linked with the corresponding patient identifiers (\texttt{subject\_id}, \texttt{hadm\_id}, \texttt{note\_id}). The extracted section titles are stored under the column \texttt{section\_name}, where the specific part under each title is stored in \texttt{section\_text} columns. Each section is assigned a unique identifier (\texttt{section\_id}) nested under the note identifier (\texttt{note\_id}). The output is organized into a DataFrame, where each row corresponds to a single section of a note, containing both metadata and textual content. An illustration of these text processing steps is described in Appendix \ref{apd:first}, Figure \ref{fig:text-integration}.

\subsubsection{Waveforms processing}\label{wv_concat}
The MIMIC-IV Waveform Database \citep{mimicwaveforms} provides continuous, high-resolution physiological signals obtained from bedside monitors during ICU stays. This resource contains the waveform recordings linked to the same patients represented in MIMIC-IV, with \texttt{subject\_id} and \texttt{study\_id} identifiers. Since the \texttt{hadm\_id} is missing this metadata, for linking them with the admission table, we connected them with the \texttt{subject\_id}.

In addition to the waveforms collection, we integrate the MIMIC-IV-ECG \citep{gow2023mimic-ecg}, an open-access collection of electrocardiogram (ECG) waveforms recorded from similar medical sources linked to MIMIC-IV patient records. 
For each patient, the ECG is stored as 12 leads and is a 10-second waveform, and includes other metadata and relating identifiers \texttt{subject\_id}, and \texttt{study\_id}.

\subsubsection{Imaging Data Processing}\label{img_concat}
The pipeline collects chest radiographs from the MIMIC-CXR \citep{johnson2019mimic-cxr} database (latest version, v2.0.0), which provides JPEG renderings of de-identified chest X-ray, Digital Imaging and Communications in Medicine (DICOMs), together with study- and image-level metadata (e.g., \texttt{subject\_id}, \texttt{study\_id}, \texttt{dicom\_id}, and view position). We associate imaging data with admissions via each \texttt{subject\_id} and merge this extracted information into a single DataFrame. 

For retrieval, radiology studies are joined to the cohorts through \texttt{subject\_id}. The corresponding radiology
reports (RR) type text (from MIMIC-IV-Note) is connected alongside each study, with the \emph{INDICATION} and \emph{FINDINGS} sections. This alignment enables users to move between structured diagnoses, sectioned clinical narratives, and imaging, using the same identifier spine across modalities.

As discussed in Section \ref{sec:note-process}, unstructured notes are handled with the note filtering option for either discharge summaries (DS) or radiology reports (RR). These RR notes are associated with the waveforms and CXR image datasets, which consist of the descriptive forms of these modalities. For the MIMIC-IV-CXR dataset, additional filtering is available for the view position of the chest X-rays (e.g., posterior-anterior (PA), anterior-posterior (AP), and lateral), and it can also retrieve the number of rotations captured for different angles. 

Apart from the CXR images, the pipeline also supports the MIMIC-IV-Echo module (v0.1) \citep{gow2023mimicecho}, which contains echocardiography reports that are linked via \texttt{subject\_id} with the central MIMIC-IV database. All of the echocardiogram image sequences appear in MIMIC-IV, which are the sequences of heart images that contain different angles and views captured through ultrasound waves. 

\subsection{Modalities Integration and Alignment}

\subsubsection{Temporal Alignment} \label{sec:temp-align}
\begin{figure}
    \centering
    \includegraphics[width=0.90\linewidth]{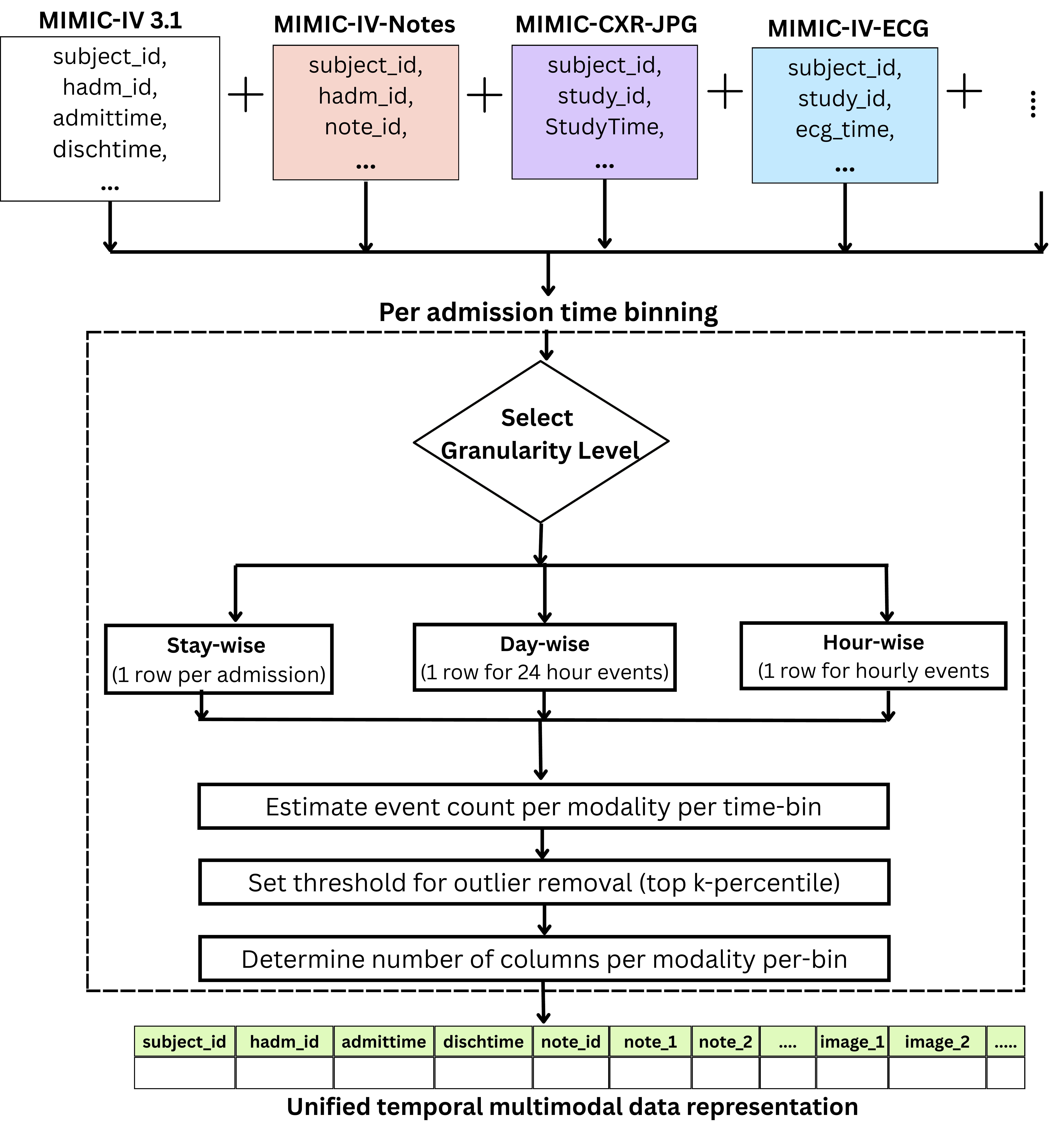}
    \caption{The overall process of temporal alignment for different modalities.}
    \label{fig:temporal}
\end{figure}

Clinical events occurring at varying frequencies have different distribution. To create a structured integration across all the multimodal events, we implemented a multi-granularity temporal binning strategy at various levels. Figure \ref{fig:temporal} illustrates the temporal alignment process. 

To preserve temporal alignment across modalities, we use \texttt{admittime} and \texttt{ dischtime} timestamps in admission records, which are associated with \texttt{hadm\_id} for each patient. These two define the timestamps when the patient was admitted and discharged from the hospital. When modalities are merged via shared identifiers (\texttt{subject\_id}, \texttt{hadm\_id}, \texttt{study\_id}), their associated timestamps are preserved in the unified DataFrame.
As patients may have multiple hospital admissions over time, our pipeline treats each admission independently by always including \texttt{hadm\_id} in all grouping keys.  

All events during the entire hospital stay ([\texttt{admittime}, \texttt{dischtime}]) are aggregated into a single row per admission. This representation captures the complete clinical trajectory for downstream tasks. We categorized the temporal alignment process into three granularities while merging modalities: stay-wise, day-wise, and hour-wise. In hour-wise binning, events are grouped into 1-hour windows from admission, creating the finest temporal resolution for real-time monitoring and early warning applications. 

A major challenge for such multimodal integration is that, within the same time window, different events for different modalities can be aligned in different cardinalities (rows). For example, within a single day, a patient may have 1 discharge summary, 4 radiology reports, 6 chest X-rays, and 2 ECG signals, all occurring at different times but falling within the same time bin (e.g., 24 hours). To handle such cases, we applied a fixed column approach where multiple events within the same time window are binned in the same row but are stored with additional columns based on the maximum number of items in each modality (e.g., if most days have 4 ECG signals, the pipeline creates additional columns for signals as signal\_1, signal\_2, and so on).

Adding extra columns increases the risk of including a large number of empty columns. To reduce this risk, the pipeline allows users to define thresholds. For example, while most days have 3-4 radiology reports, there may be rare outlier days with more than 16 reports. The user can set this threshold and remove the rows with instances beyond the $k$th percentile. This ensures that $K$\% of time windows are fully represented without truncation, while avoiding the creation of many empty columns.

\subsubsection{Raw Data Integration}\label{sec:data-connect}

To make integration scalable and memory-efficient, especially for waveform (ECG/physiology) assets and large imaging (CXR and echo), the pipeline does not embed raw images or signal arrays in the integrated table. Instead, it stores the resolved \texttt{file\_path} pointers for each modality and record, constructed deterministically from the dataset metadata (e.g., file lists, view positions, and study timestamps) and the directory conventions used by PhysioNet and local directories. Users can access the images or signals by using the \texttt{file\_path} or by using the provided addresses for downstream analysis. For text, \texttt{note\_id} is used, and \texttt{study\_ids} is used to bridge images and signal type modalities. All these merged modalities can be stored as CSV files for downstream tasks. Figure \ref{fig:sample} in Appendix \ref{apd:first} presents an example of multimodal integration. 

\subsubsection{Handling Missing Values}

The pipeline integrates all available data within the specified time interval, and no rows are removed due to the absence of any modalities. 
In the final integrated datasets, the rows with missing modalities are filled with null values (for example, the corresponding cell to a missing \texttt{file\_path} for an X-ray image is filled with null).

For numeric time-series, such as lab vitals, we bin measurements into fixed intervals and attach per-bin presence indicators. Users can choose the imputation method, such as forward filling, mean, or median imputation. Medications are encoded as dosage-over-time signals and procedures as on/off events without numeric imputation.

\subsection{Embedding Representation}
As discussed earlier, in addition to generating raw multimodal data, the pipeline provides the option to convert the merged data into consolidated embeddings, which are stored in distinct columns as \texttt{text\_embed}, \texttt{signal\_embed}, and \texttt{img\_embed}. These embeddings offer users a flexible option for implementing pre-processed embedded data. A similar process to what was discussed in Section \ref{sec:temp-align} is used for embeddings, where each \texttt{subject\_id} is connected via the corresponding \texttt{even\_time}. The pipeline includes several popular embedding models as built-in options, but users can also use their preferred models. 

For text embeddings, the pipeline integrates \texttt{Bio\_ClinicalBERT} \citep{bioclinicalbert}, a transformer-based model pre-trained on biomedical corpora, as a baseline.  It obtains the embeddings of the text column from the unstructured MIMIC-IV-Note data. Other advanced embedding models, including \texttt{ModernBERT} \citep{modernbert}, \texttt{BioSimCSE-BioLinkBERT} \citep{kanakarajan2022biosimcse}, and \texttt{GatorTron} \citep{yang2022gatortron}, are also available in the pipeline. For the larger notes, the pipeline chops the tokens into 512 tokens and overlaps them by 64 strides. After encoding with the \texttt{BERT} \citep{devlin2019bert} model, the resulting hidden states are aggregated using mean pooling across these tokens. The chunks for the corresponding \texttt{note\_id} are averaged and presented in a 768-dimensional representation.

For waveform data, physiological raw signals (e.g., ECG signals) are preprocessed into uniform-length tensors, with optional resampling applied to handle varying frequencies. Here, the pipeline supports using the classic \texttt{Resnet1d101} \citep{he2016resnet} as a backbone with the pretrained \texttt{torch\_ecg} \citep{torch-ecg-ptbxl}, which is pre-trained on a large ECG dataset (PTB-XL) \citep{wagner2020ptb}. Users can also select between models such as \texttt{ConvNeXt}\citep{liu2022convnet}, \texttt{EfficientNetV2-S} \citep{tan2021efficientnetv2}, and the self-supervised transformer model \texttt{DINOv2} \citep{oquab2023dinov2} for waveform embeddings. As with text data, waveforms are merged with \texttt{study\_id}, thereby aligning \texttt{subject\_id}. 

For imaging data (e.g., chest X-rays and echocardiograms), raw pixel arrays are preprocessed into normalized tensors following standard image-processing practices. Metadata, such as view position or study information, is retained to preserve context, while the images are converted into embeddings and stored as compact column entries. For the image embeddings, the pipeline uses the \texttt{study\_id} and \texttt{dicom\_id} that are matched with the \texttt{subject\_id}. The pipeline supports a variety of vision backbones, ranging from lightweight models like \texttt{EfficientNet-B0} \citep{effnet}, \texttt{ResNet101}\citep{he2016resnet}, to deeper architectures, such as \texttt{DenseNet121} \citep{huang2017densenet}. Table \ref{tab:mm-embed} in Appendix \ref{apd:first} illustrates a snapshot example, showing how the embeddings are integrated. 



Our pipeline provides a customized location for downloaded files that can accommodate larger data files, such as notes or images. For that, the pipeline allows users to choose between storing raw data locally or embedding it, and to configure the system to download files from the Physionet website using their username and password. This flexible design provides greater opportunities to experiment with the sample datasets. 



\subsection{Python Package}
We present our pipeline in the form of a Python package, \texttt{MIMICEmbedding}\footnote{\url{https://pypi.org/project/MIMICEmbedding/0.1/}} that includes the functionalities as ICD-based cohort generation to generate embeddings, when required, the \texttt{cohort()} function matches with the and returns the tensor and embeddings for the associated \texttt{subject\_ids} for waveforms, X-ray images, and 2-D echocardiograms. There is another feature that allows the rotation of the images to be selected for the same cohort. Finally, the \texttt{combine()} function combines all the modalities on the same Dataframe.

This pipeline also offers another lightweight package, \texttt{MIMICSectionizer}\footnote{\url{https://pypi.org/project/MIMICSectionizer/}} that performs the regex-based segmentation of unstructured notes into distinct sections (e.g., History of Present Illness or Findings). For cohort-based searching, a function \texttt{search()} (available in both packages) is developed that performs ICD-9- or ICD-10-based cohort searches, linking across modalities, along with text embeddings. The package includes APIs that help users execute each part independently without additional coding effort. 
The packages can be installed locally using the command \texttt{`pip install -e'}.

\begin{table*}[htb]
\centering
\small
\caption{Comparison among MIMIC Data processing pipelines with our multimodal pipeline}
\label{tab:compare}
\resizebox{\textwidth}{!}{%
\begin{tabular}{l |l |c |c |c| c | c | c}
\toprule
\textbf{Features} & \textbf{MIMIC-Extract} & \textbf{FIDDLE} & \textbf{COP-E-CAT} & \textbf{HAIM-MIMIC-MM} &\textbf{Gupta et el.}&\textbf{Quick-MIMIC}& \textbf{Our Pipeline} \\
\midrule
Total modalities   & 1  & 1 & 1 & 4 & 2 & 2 & 5 \\
Structured  &  \ding{51} & \ding{51} & \ding{51} & \ding{51} & \ding{51} & \ding{51} & \ding{51} \\
Clinical notes    & \ding{55}  & \ding{55} & \ding{55} & \ding{51} & \ding{55} & \ding{51} & \ding{51} \\
CXR-images  & \ding{55}  & \ding{55} & \ding{55} & \ding{51} & \ding{55} & \ding{55} & \ding{51} \\
ECG-signals & \ding{55}  & \ding{55} & \ding{55} & \ding{51} & \ding{55} & \ding{55} & \ding{51} \\
Echocardiograms    & \ding{55}  & \ding{55} & \ding{55} & \ding{55} & \ding{55} & \ding{55} & \ding{51} \\
Physical waveforms  & \ding{55}  & \ding{55} & \ding{55} & \ding{51} & \ding{55} & \ding{55} & \ding{51} \\
Customizable cohort selection & \ding{55}  & \ding{55} & \ding{55} & \ding{55} & \ding{55} & \ding{55} & \ding{51} \\
Text sectionizer   & \ding{55}  & \ding{55} & \ding{55} & \ding{55} & \ding{55} & \ding{55} & \ding{51} \\
Pre-computed embeddings  & \ding{55}  & \ding{55} & \ding{55} & \ding{51}  & \ding{55} & \ding{55} & \ding{51}\\
Embeddings flexibility  & \ding{55}  & \ding{55} & \ding{55} & Fixed & \ding{55} & \ding{55} & Customizable\\
Image filtering  & \ding{55}  & \ding{55} & \ding{55} & \ding{55} & \ding{55} & \ding{55} & \ding{51}\\
\bottomrule
\end{tabular}
}
\end{table*}

\section{Experiments} 
\label{sec:exp}

As our study primarily aims to offer a user-friendly pipeline for pre-processing MIMIC data, we report a series of analyses and demonstrative experiments to show the versatility and utility of the pipeline.

\paragraph{Feature support vs. other MIMIC tools.} We compare our pipeline with six other MIMIC pipelines to demonstrate the overall similarities and differences they have. The studied pipelines include \texttt{MIMIC-Extract} \citep{wang2020mimicextract}, \texttt{FIDDLE} \citep{tang2020fiddle}, \texttt{COP-E-CAT} \citep{mandyam2021cop}, \texttt{HAIM-MIMIC-MM} \citep{soenksen2022haim}, \citet{gupta2022extensive}, and \texttt{Quick-MIMIC} \citep{dou2024quickmimic}. Table \ref{tab:compare} highlights the features that our pipeline supports compared to other recent works. 

Apart from \texttt{HAIM-MIMIC-MM}, other works are only focused on single or just two modalities. 
\texttt{HAIM-MIMIC-MM} provides pre-computed embeddings for some modalities, but those are fixed and cannot be customized to meet specific research needs. 


\paragraph{Demonstrative Case Studies.}
To demonstrate the functionality of our multimodal pipeline, we coupled the pipeline with a series of diverse downstream tasks, showing its wide usage. In particular, this pipeline targets two popular predictive tasks, including predicting, i) in-hospital mortality, and ii) length of stay (in ICU) greater than three days. We target three types of cohorts, based on the patients' diagnosis codes. These diagnosis codes include, i) Atrial Fibrillation (ICD-9: 42731), ii) Coronary Artery Disease (414*, I25*), and iii)
Sepsis-related heart diseases (`038', `99591', `99592', `78552', `A40', `A41', `428', `410', `I50', `I21'). We present additional details about the cohort extraction in Appendix \ref{apd:exp-setup}.



Table \ref{tab:mm_result} represents some exemplary performance of our pipeline across different settings. While there is less focus on the performance values, the diverse settings used in the experiments showcase the ability of our pipeline to support various configurations of downstream tasks. 

\begin{table*}[ht]
\centering
\small
\caption{Multimodal prediction analysis for customized cohorts demonstrating the versatility of our pipeline across different settings.}
\label{tab:mm_result}
\resizebox{\textwidth}{!}{
\begin{tabular}{l l l c c c c}
\toprule
\textbf{Prediction Tasks} & \textbf{Downstream Model} & \textbf{Cohort} & \textbf{Modalities} & \textbf{Accuracy} & \textbf{AUROC} & \textbf{AUPRC} \\
\midrule
\multirow{12}{*}{In-Hospital Mortality} &
1D-CNN + MLP    &  Atrial Fibriliation & Structured + Waveforms (ECG Signals)        & 0.61 & 0.68  & 0.65 \\
& Bi-LSTM + MLP  &   Atrial Fibriliation       & Structured + Waveforms (ECG Signals)        & 0.65 & 0.69  & 0.69 \\
& Hybrid (Transformer+CNN) & Atrial Fibriliation & Structured + Waveforms (ECG Signals)        & 0.64 & 0.68  & 0.60 \\
\cmidrule(lr){2-7}
& DenseNet121 + MLP   &   Atrial Fibriliation  & Structured + CXR images    & 0.90 & 0.96  & 0.95 \\
&ResNet18 + MLP      &  Atrial Fibriliation   & Structured + CXR images    & 0.90 & 0.98  & 0.97 \\
& EfficientNet + MLP   &  Atrial Fibriliation  & Structured + CXR images   & 0.89 & 0.97  & 0.97 \\
\cmidrule(lr){2-7}
& Gradient Boosting   &  Coronary Artery Diseases  & Structured (w/ Time-series) + CXR images   & 0.69 & 0.83  & 0.79 \\
& XGBoost  &   Coronary Artery Diseases   & Structured (w/ Time-series) + CXR images    & 0.69 & 0.97  & 0.79 \\
& CatBoost  & Coronary Artery Diseases  & Structured (w/ Time-series) + CXR images   & 0.69 & 0.81  & 0.73 \\

\cmidrule(lr){2-7}
& Gradient Boosting   &  Coronary Artery Diseases  & Structured + Notes + Waveforms (ECG Signals)    & 0.69 & 0.83  & 0.79 \\
& XGBoost  &   Coronary Artery Diseases   & Structured + Notes + Waveforms (ECG Signals)    & 0.69 & 0.98  & 0.79 \\
& Transformers  & Coronary Artery Diseases  & Structured + Notes + Waveforms (ECG Signals)    & 0.59 & 0.63  & 0.70 \\
\cmidrule(lr){1-7}
    \multirow{3}{*}{Length of Stay $>$ 3}
      & DenseNet121 + MLP &  Atrial Fibriliation & Structured + CXR images & 0.90 & 0.96 & 0.96 \\
      & EfficientNet + MLP &Atrial Fibriliation & Structured + CXR images & 0.92 & 0.96 & 0.96 \\
      & ResNet18 + MLP & Atrial Fibriliation & Structured + CXR images & 0.90 & 0.95 & 0.95 \\
      \cmidrule(lr){2-7}
     & DenseNet121 + MLP  & Sepsis-related heart disease & Structured + CXR images    & 0.68 & 0.71  & 0.77 \\
& ResNet18 + MLP & Sepsis-related heart disease & Structured + CXR images    & 0.59 & 0.67  & 0.74 \\
& EfficientNet + MLP   &  Sepsis-related heart disease  & Structured + CXR images   & 0.63 & 0.61  & 0.69 \\
\bottomrule
\end{tabular}
}
\end{table*}

\paragraph{Run times.} We also report the run times for extracting raw integrated datasets for several combinations of modalities using some commodity hardware (a desktop PC with Intel Core i7 CPU and 16 GB of DDR4 RAM, running Windows 11). Table \ref{tab:runtime} shows the total running (clock) time for the combined multimodal datasets. When text data are included, integration time increases.  , with text+CXR(100) requiring approximately 2,430 seconds (40 minutes) and text+Wave(100) requiring 2,479 seconds (41 minutes). On the other hand, integrating waveforms alone is nearly instantaneous, as a cohort of 50 waveform records required only 0.31 seconds, while 100 waveform records were still completed in under a second. 


\begin{table}[ht]
  \centering
  \small
  \caption{Runtime (in seconds) for different combinations of MIMIC-IV integrated modalities.  $n$ is the sample size of the data.}
  \resizebox{\columnwidth}{!}{
    \begin{tabular}{l|l}
      \toprule
      \textbf{Integrated Modalities} & \textbf{Runtime} \\ \toprule
      Notes + X-ray (n:100) & 2,430.44  \\ \hline
      Notes + Waveforms (n:100) & 2,479.06 \\ \hline
      Waveforms (n:50) & 0.31 \\ \hline
      X-ray (n:100) + Waveforms (n:50) & 25.78 \\ \hline
      X-ray  (n:100) + Waveforms (n:100)  & 26.70 \\\hline
      Structured + Notes + Waveforms (n:200) & 300 \\\hline
      Structured + Notes + Waveforms (n:200) + X-ray (n:200) + 2D Echo (n:200) & 2,735.85 \\\bottomrule
    \end{tabular}%
  }
  \label{tab:runtime}
\end{table}
\section{Discussion}

The proposed multimodal pipeline addresses a critical gap in the use of the MIMIC-IV dataset: the lack of a unified framework for integrating structured EHR data, unstructured clinical notes, waveforms, X-ray images, and 2D-Echocardiograms modalities. By systematically linking datasets through shared identifiers (\texttt{subject\_id}, \texttt{note\_id}, and \texttt{study\_id}) and leveraging metadata to generate file paths for large imaging and waveform assets, our approach provides a scalable and memory-efficient integration strategy. This design enables researchers to access multimodal patient data flexibly, eliminating the heavy burden of redundant preprocessing or manual alignment.

A major goal of the pipeline is to save users' time and reduce technical overhead. Traditionally, researchers and practitioners (especially those new to the field) have to invest considerable effort in understanding the internal database schema, the distinct storage conventions of each modality, and the unique identifier mappings needed to connect them. Our pipeline can significantly reduce this overhead, thereby saving time and eliminating redundancy in the data cleaning steps. 

The pipeline builds upon and extends previous attempts to extract MIMIC data, specifically aiming to integrate all best practices for handling EHR data, including clinical grouping of codes (phenotypes), temporal alignment, handling missing values, and removal of outliers (including physiologically implausible values). The pipeline also includes standard methods for evaluation of the downstream tasks, including discriminative performance, bias evaluation, and calibration. 

The runtime evaluation demonstrates that even when integrating large combinations of text, images, and waveform data, the process completes within minutes rather than hours or days that manual preprocessing would typically require \citep{gupta2022extensive, wang2020mimicextract}. For smaller subsets or exploratory analyses, the integration can be completed almost instantaneously. These results underscore the pipeline’s potential to democratize access to multimodal EHR research by enabling new and early-stage researchers to focus on methodological and scientific questions rather than technical infrastructure. 

Beyond efficiency, the framework also enhances reproducibility and comparability across studies. By standardizing preprocessing, ICD-based cohort selection, notes preprocessing, and cross-modality integration, the pipeline reduces the variability and inconsistency introduced while structuring diverse data modalities. This is especially important in the medical AI domain, where reproducibility remains a persistent challenge and inconsistent preprocessing can significantly influence downstream results. 

\paragraph{Limitations}

Our work has several limitations. For temporal alignment, the pipeline uses the \texttt{admittime} to \texttt{dischtime} window, which includes the inpatient entries while excluding pre-admission entries for any event. This design choice is consistent with the approach taken by others \citep{soenksen2022haim}, and is made to handle the selection bias \citep{khaled2025mimictemporal}, which reduces the likelihood of selecting incomplete admissions caused by imbalanced multimodal data availability across patients. The pipeline, however, documents the alternative ways that the users can implement the alignments. Relatedly, instead of a complete-case approach for integration (where only records with all modalities being available are picked), we used a temporal window based on a specific time interval, which can result in many null values for missing modalities. 

Furthermore, the clinical note sectionizer is currently rule-based (regex), which is computationally efficient, but may fail to parse notes with non-standard templates or typos. To maximize relevance to MIMIC and the utility of the proposed tool, the pipeline is tightly coupled to the MIMIC-IV schema and is not a general-purpose EHR tool. 

\acks{Our work was supported by NSF award 2443639, and NIH awards, P20GM103446, and
U54GM104941. }
\bibliography{references}

\appendix

\section{Additional Figures and Results}\label{apd:first}

\begin{figure*}[ht!]
    \centering
    \includegraphics[width=0.85\linewidth]{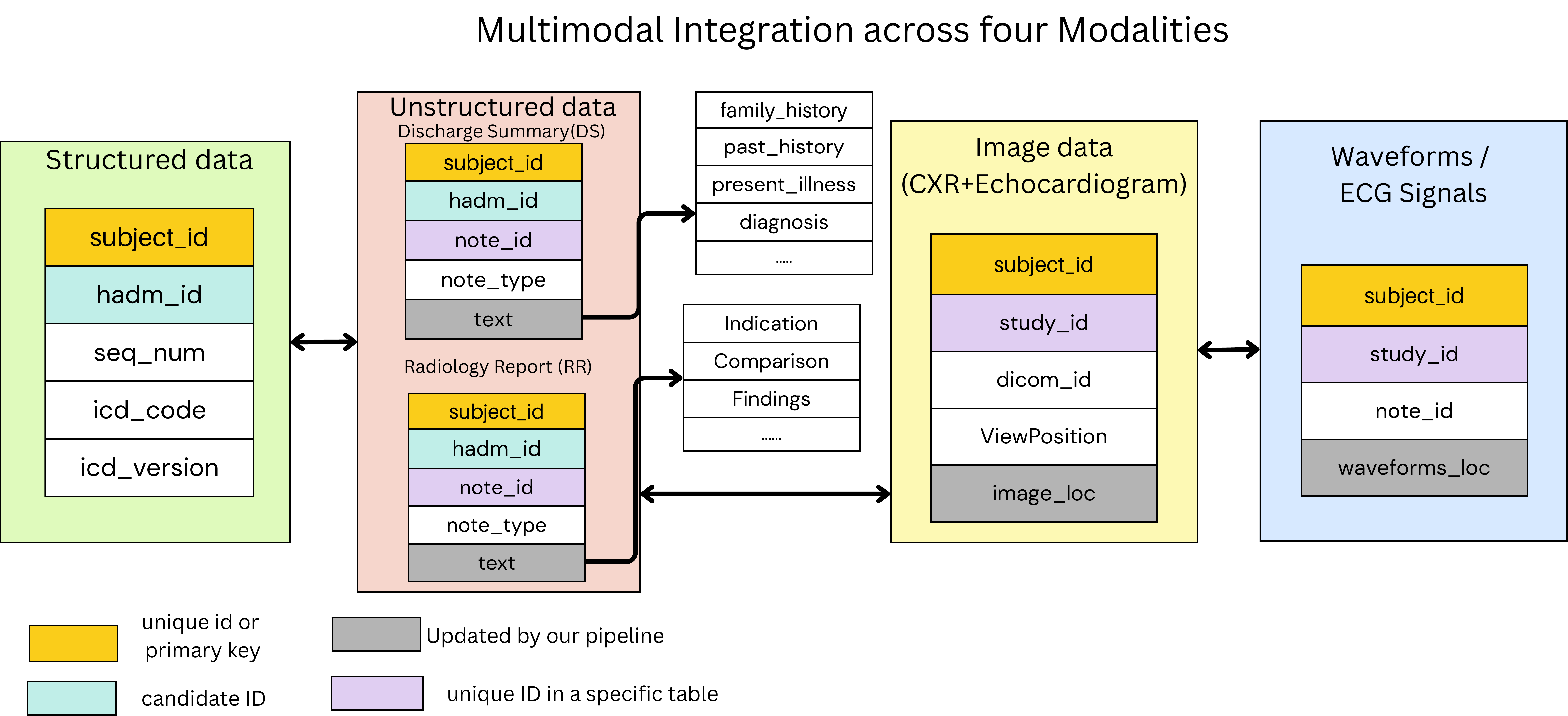}
    \caption{A sample demonstration of how the database tables are integrated based on the unique identifiers.}
    \label{fig:sample}
\end{figure*}

\begin{figure*}[ht!]
    \centering
    \includegraphics[width=0.65\linewidth]{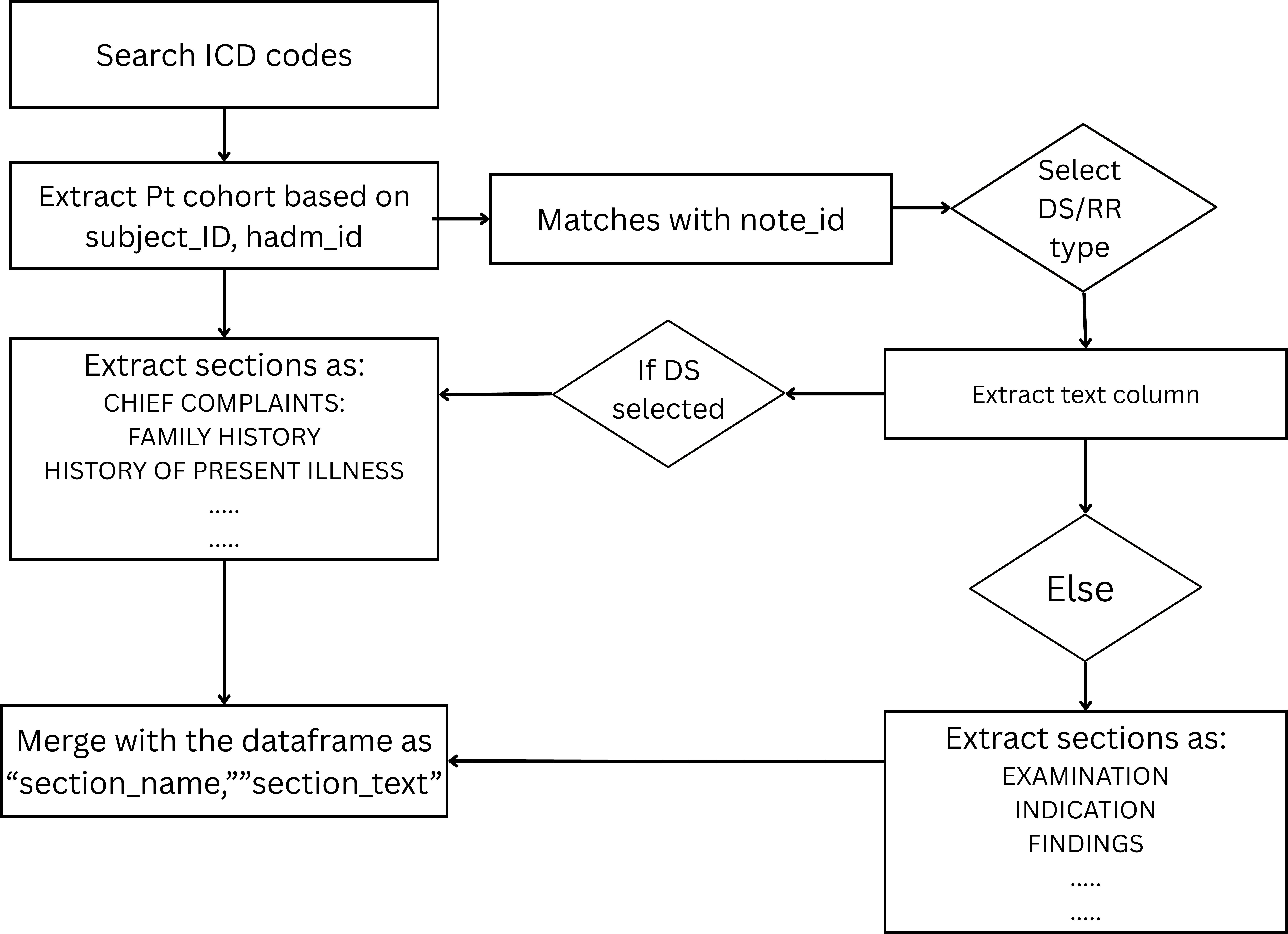}
    \caption{An overview of the text extraction process from MIMIC-IV clinical notes.}
    \label{fig:text-integration}
\end{figure*}
We have also conducted another cohort-based analysis, with the Acute Kidney Injury (AKI) (ICD Codes: N179, 5849) for predicting ICU length of Stay for 3 and 7 days, with a set of structured features and CXR images. Table \ref{tab:aki-los} shows the performance of the LOS prediction for the AKI patient cohort. 

\begin{table}[ht]
  \centering
  \caption{Results by task with identical model sets per task.}
  \label{tab:aki-los}
  \resizebox{\columnwidth}{!}{%
  \begin{tabular}{l l l c c c}
    \toprule
    \textbf{Prediction Task} & \textbf{Model} & \textbf{Modalities} & \textbf{Accuracy} & \textbf{AUROC} & \textbf{AUPRC} \\
    \midrule
    \multirow{2}{*}{LOS $>$ 3} 
      & DenseNet121 + MLP & CXR images + Structured & 0.655 & 0.879 & 0.923 \\
      & EfficientNet + MLP& CXR images + Structured & 0.651 & 0.828 & 0.873 \\
    \cmidrule(lr){1-6}
    \multirow{2}{*}{LOS $>$ 7}
      & DenseNet121 + MLP & CXR images + Structured & 0.605 & 0.705 & 0.616 \\
      & EfficientNet + MLP& CXR images + Structured & 0.619 & 0.783 & 0.644 \\
    \bottomrule
  \end{tabular}%
  }
\end{table}

\begin{table}[ht]
  \centering
  \small
  \caption{Example of Multimodal Embedding Representation}
  \resizebox{\columnwidth}{!}{
    \begin{tabular}{l|l|l|p{4cm}|p{4cm}|l}
      \toprule
      \textbf{subject\_id} & \textbf{study\_id} & \textbf{note\_id} & \textbf{text\_embed} & \textbf{img\_embed} & \textbf{\dots}\\ \toprule
      10313763 & 51527697 & 10313763-RR-37 & [0.005195111967623234, -0.03489554673433304, -\dots & [2.2592575987800956e-06, 5.175691330805421e-05, \dots & \dots\\ \hline
      10404109 & 52328790 & 10404109-RR-12 & [0.009105992503464222,   -0.025197483599185944, \dots & [2.9136190278222784e-06, 0.0001181846964755095, \dots & \dots\\ \hline
      12133670 & 51759637 & 12133670-RR-13 & [0.0075526065193116665, -0.017450641840696335, \dots & [3.84774921258213e-06, 0.0001566458522574976, \dots & \dots\\ \hline
      \dots & \dots & \dots & \dots & \dots & \dots\\
    \bottomrule
    \end{tabular}%
  }
  \label{tab:mm-embed}
\end{table}

\begin{table*}[h!]
\centering
\small
\caption{Comparative Analysis for the Performance Evaluation of Downstreaming tasks (Prediction)}
\label{tab:compare-cases}
\resizebox{\textwidth}{!}{%
\begin{tabular}{l |p{3cm} |p{3cm} |p{3cm} |p{3cm}| p{3cm} | p{3cm} | p{3cm}}
\toprule
\textbf{Features} & \textbf{MIMIC-Extract}\citep{wang2020mimicextract} & \textbf{FIDDLE}\citep{tang2020fiddle} & \textbf{COP-E-CAT} \citep{mandyam2021cop} & \textbf{HAIM-MIMIC-MM }\citep{soenksen2022haim}&\textbf{\citet{gupta2022extensive}}&\textbf{Quick-MIMIC} \citep{dou2024quickmimic} & \textbf{Our Pipeline} \\
\midrule
MIMIC Version  & MIMIC III  & MIMIC III & MIMIC IV & MIMIC IV & MIMIC IV & MIMIC IV 2.2 & MIMIC IV 3.1 \\\hline
Modalities Combination  & Structured (time-series)  & Structured & Structured (time-series) & Tabular, time-series, text, and images & Structured (time-series) & Structured, text & Tabular, time-series, text, signals, and images \\\hline
Similar Downstream tasks &  Mortality, LOS & Mortality & Intervention prediction & Prediction ( Mortality, LOS) & Prediction (Mortality, LOS) & \ding{55} & Cohort-based prediction (Mortality, LOS) \\\hline
Cohort used  & ICU admitted all patients & ICU
visits & ICU admitted all patients & Hospital stays (6485 unique) & Patient with chronic kidney diseases (CKD) & \ding{55} & Disease-specific (AF, AKI) \\\hline
Sample size  & 53,423 patients (ICU)  & 23,620 & Customizable ($\geq 18$ yrs old) & 45,000+ & Not specified & 50,000 inpatient cases & Customizable \\\hline
Variables  & Demographic variables, labevents, timestamps  & Demographic variables, labevents & Demographic variables, labevents& Demographic variables, notes, timestamps, X-ray images & Demographic variables, labevents, timestamps & Demographic variables, labevents, timestamps, notes & Demographic variables, notes, timestamps, X-ray images, waveforms \\\hline
Model Architecture & LR, RF, GRU-D  & CNN, LSTM &  Markov decision process (MDP) & XGBoost (embeddings) & LR, RF, LSTM, TCN & \ding{55} & signals: BiLSTM, CNN, images: EfficientNet, ResNet, DenseNet \\\hline
\multirow{2}{*}{AUROC} & 85.6-87.6 (Mortality)  & 82.1-91.6 (Mortality) & 0.86 (Prediction) & $\approx .93$ (Mortality) & .67 -.85 (Mortality) & \ding{55} & .68 -.97 (Mortality) \\
\cmidrule(lr){2-8}
& 71-73 (LOS)  & \ding{55} & \ding{55} & $\approx .85$ (LOS) & .64 -.80 (LOS) & \ding{55} & .61 -.88 (LOS) \\
\bottomrule
\end{tabular}
}
\end{table*}

\section{Experimental Setup}
\label{apd:exp-setup}
For each patient, we randomly sampled 1,000 chest X-ray (CXR) images from the MIMIC-CXR-JPG v2.1.0 collection, ensuring that each image is linked to the admission record through the unique subject and study identifiers. Similar to prior work \cite{hamed_fayyaz_interoperable_2024, pmlr-v298-piya25a,adiba2025bias,gupta_reliable_2024,Mottalib25,adiba2025reveal}, along with the images, admission-level tabular attributes are used, where we only considered a subset of sample features (e.g., `age', `insurance type', `race'), which are extracted by matching with the image metadata using a unique identifier. For predicting in-hospital mortality, the \texttt{hospital\_expire\_flag} column from `admissions.csv' is used as the target outcome. 

We implemented three hybrid multimodal architectures, DenseNet121, ResNet18, and EfficientNetv2 through the \citet{effnet} instances, combining convolutional neural networks (CNNs) for image feature extraction with a multilayer perceptron (MLP) for structured feature integration. For waveforms, we randomly sampled 500 ECG signals from MIMIC-IV-ECG v1.0  with other tabular features to predict in-hospital mortalities, implementing three hybrid models.

To handle imbalanced scenarios in the experiments 
we applied the SMOTE (Synthetic Minority Oversampling Technique) oversampling approach, which assigns extra weight to the minority classes. 
\end{document}